\documentclass{ifacconf}

\usepackage{xcolor}
\usepackage{graphicx}      
\usepackage{natbib}        
\usepackage{amsmath}
\usepackage{float}
\usepackage{caption}
\usepackage{subcaption}
\usepackage{svg}
\usepackage{algorithm}
\usepackage{algpseudocode}
\usepackage{multirow}
\usepackage{flushend}
\DeclareMathOperator*{\argminB}{argmin}

\usepackage[nolist]{acronym}
\begin{acronym}
  \acro{GP}{Gaussian process}
  \acroplural{GP}[GPs]{Gaussian processes}
\end{acronym}

\begin{document}
\begin{frontmatter}

\title{Ensemble Gaussian Processes \\
for Adaptive Autonomous Driving on Multi-friction Surfaces}

\author[1]{Tomáš Nagy,} 
\author[1]{Ahmad Amine,} 
\author[2]{Truong X. Nghiem,}
\author[3]{Ugo Rosolia,}
\author[1]{Zirui Zang,}
\author[1]{Rahul Mangharam}

\address[1]{University of Pennsylvania, Philadelphia, PA 19104, USA \{nagytom, aminea, zzang, rahulm\}@seas.upenn.edu}
\address[2]{School of Informatics, Computing, and Cyber Systems; Northern Arizona University (Truong.Nghiem@nau.edu)}
\address[3]{Research Scientist, Amazon, 22 Rue Edward Steichen,
2540 Luxembourg (ugo.rosolia)@gmail.com}

\begin{abstract}                
Driving under varying road conditions is challenging, especially for autonomous vehicles that must adapt in real-time to changes in the environment, e.g., rain, snow, etc.
It is difficult to apply offline learning-based methods in these time-varying settings, as the controller should be trained on datasets representing all conditions it might encounter in the future.
While online learning may adapt a model from real-time data, its convergence is often too slow for fast varying road conditions.
We study this problem in autonomous racing, where driving at the limits of handling under varying road conditions is required for winning races.
We propose a computationally-efficient approach that leverages an ensemble of Gaussian processes (GPs) to generalize and adapt pre-trained GPs to unseen conditions.
Each GP is trained on driving data with a different road surface friction.
A time-varying convex 
combination of these GPs is used within a model predictive control (MPC) framework, where the model weights are adapted online to the current road condition based on real-time data.
The predictive variance 
of the ensemble Gaussian process (EGP) model allows the controller to account for prediction uncertainty and enables safe autonomous driving.
Extensive simulations of a full scale autonomous car 
demonstrated the effectiveness of our proposed EGP-MPC method for providing good tracking performance 
in varying road conditions and the ability to generalize to unknown maps.
\end{abstract}


\end{frontmatter}


\section{Introduction}
Driving under varying road conditions is a challenging task, even for experienced drivers. Safe driving requires constant monitoring of road conditions and adjusting the driving strategy accordingly~\cite{laurense2017path}. To deploy driver-assist technologies and autonomous vehicles in the real-world, we must ensure that these systems can drive safely under all conditions, which are often time-changing while driving. This fact motivates the use of learning-based control approaches that can adapt during execution~\cite{rosolia2018data, hewing2020learning}.

Several learning-based approaches for autonomous driving 
have been developed in recent years \cite{Betz2022_RacingSurvey}. In \cite{jain_bayesrace_2020}, vehicle kinematics are corrected by identifying residual errors with \acp{GP}. A nonlinear model predictive controller then uses the corrected model to track the desired trajectory. 
\cite{rosolia_learning_2019} compute a local linear regressor to model the vehicle dynamics and then iteratively construct safe sets used as terminal constraints in an MPC design. In \cite{le_receding_2021}, the conditional differential entropy of a \ac{GP} that models the vehicle dynamics was added to a model predictive controller to push the controller toward the most informative states while maintaining stability. \cite{hewing2018cautious} use a sparse GP approximation with simplified chance constraints to design a real-time controller that outperforms standard nonlinear model predictive control (NMPC) for autonomous miniature race cars. While these approaches can control their systems using a learned model, they assume operating conditions are not changed and are consistent throughout training and testing. In this paper, we show that using one model through different driving operating conditions does not always suffice, and we present an algorithm capable of handling changing conditions. A similar attempt to model and adapt to changing conditions is done in \cite{vaskov_SNMPC_2022}, where a learned \ac{GP} of friction is used to formulate a chance constraint stochastic nonlinear model predictive controller. In this paper, we capture model uncertainties as we do not assume prior knowledge of system dynamics but rather learn these dynamics with \acp{GP} too.

The contribution of this paper is threefold. First, we present an ensemble of \acp{GP} as a time-varying system model that can be used for control. Our approach differs from the work done in \cite{jain_bayesrace_2020, hewing2018cautious,rodriguez_learning_2021}, as we leverage a library of \acp{GP} to achieve lower prediction error. We show that in the presence of changing environments, a single \ac{GP} is unable to achieve the best estimate, while our method can compute accurate predictions. Second, we provide a weight smoothing algorithm for ensembling the library of pre-trained \acp{GP}. Finally, we demonstrate how these algorithms can be used to design a model predictive controller for autonomous driving in varying surface conditions.

The paper is organized as follows. In Section~\ref{sec:problem_formulation}, we introduce the problem formulation. Section~\ref{section:EGP} describes the proposed GP ensemble strategy. The control design methodology is presented in Section~\ref{sec:MPC}. Finally, simulation results are discussed in Section~\ref{sec:results}.

\section{Problem formulation}
\label{sec:problem_formulation}

\begin{figure*}[t]
    \centering
    \vspace{-10pt}
    \includegraphics[width=1.0\textwidth]{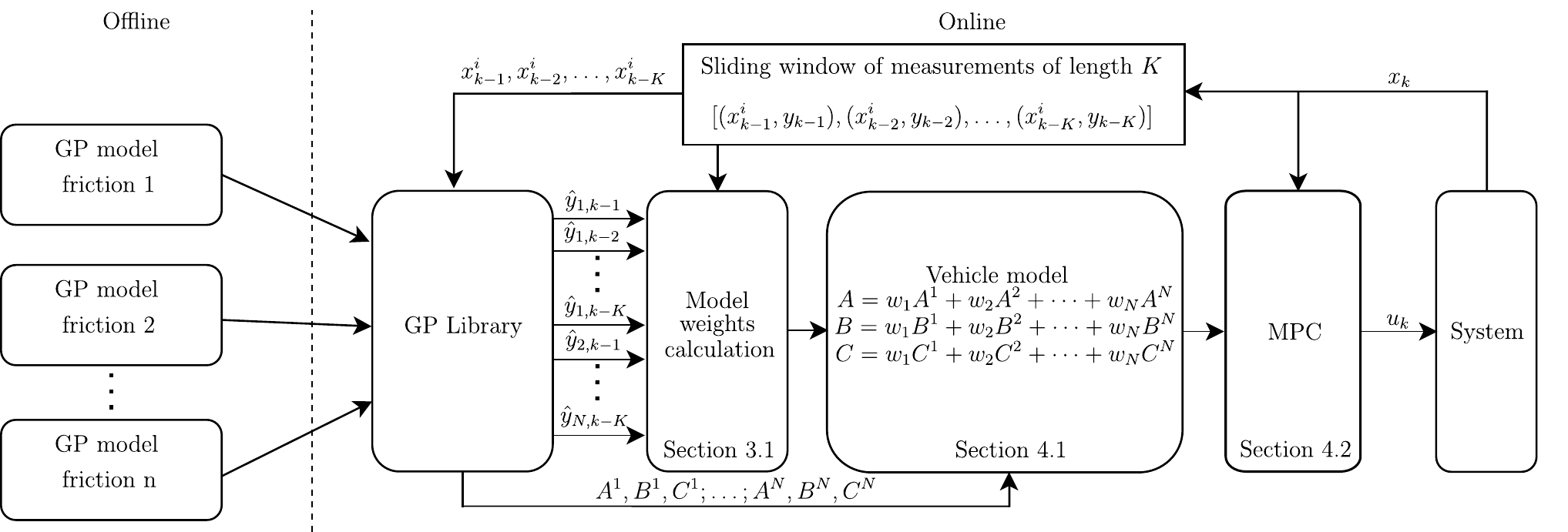}
    \caption{System pipeline: At every time step, we have a vector of past observation (inputs and outputs for the \acp{GP}) $[(x^i_{k-1}, y_{k-1}), \dots, (x^i_{k-K}, y_{k-K})]$. We use inputs to the \acp{GP} $[x^i_{k-1}, \dots, x^i_{k-K}]$ and a library of \acp{GP} to predict outputs $\hat{y}_{k}$. Then we use measured outputs $y$ and predicted outputs $\hat{y}$ to calculate the combination of models (weights $w$) that best represents the real behavior over the history of observations. We use weights $w$ and the library of \ac{GP} models to create ensemble matrices $\hat{A}$, $\hat{B}$, and $\hat{C}$ which we use for the MPC.}
    \label{fig:system_pipeline}
\end{figure*}

We consider the following 
nonlinear dynamic system 
\begin{equation}
    \label{eq:system-dynamics}
    x_{k+1} = f(x_k, u_k; \theta^n) + \epsilon_k
\end{equation}
where, at time $k$ and mode $n$, $x_k$ is the system state, 
$u_k$ is the control input, 
$\epsilon_k$ is the noise, and $\theta^n$ is the set of system parameters. 
The system is subject to 
state and input constraints
\begin{equation*}\label{eq:cnstr}
    x_k \in \mathcal{X} \text{ and } u_k \in \mathcal{U},
\end{equation*}
for all time step $k \in \{0, 1, \ldots\}$.
The parameters $\theta^n$ are assumed to be unknown at runtime. 

For controlling the system \eqref{eq:system-dynamics}, we aim to learn a data-driven model of it.
This is particularly challenging as the system parameters $\theta^n$ are both unknown and varying over time.
To tackle this challenge, we propose an ensemble data-driven modeling approach that combines several estimated models of the system, which are trained offline from system data obtained under different operating conditions, and adapts their combination in real time.

Suppose that we can collect experimental data of the system in $N$ different operating conditions with unique sets of parameters $\theta^1,\dots,\theta^N$.
We do not assume that the parameters are known, however, in each controlled operating condition, the parameters are constant so that the collected data is consistent with the specific operating condition. 
For each operating condition with parameters $\theta^n$, 
we collect time series of 
input-state pairs of the system 
in the form
\begin{equation}
    \label{eq:data}
    \mathcal{D}^n = \left\{
        (\hat{u}^n_1, \hat{x}^n_1),\dots,(\hat{u}^n_m, \hat{x}^n_m)
    \right\}\text.
\end{equation}

In the context of autonomous driving in this paper, we consider the road friction as the system parameter $\theta^n$ as it can greatly affect a vehicle's dynamics and its driving performance, but is often unknown in real time.
Furthermore, in practical driving, road friction varies due to various factors, such as the type of road and the weather condition.
Our proposed ensemble learning approach builds a library of offline models trained for different friction surfaces then combines them to adapt to the actual friction under the real time driving conditions.
We then leverage the adaptive ensemble model in a model predictive control framework to drive the vehicle autonomously.
The overall pipeline of our approach is illustrated in Figure~\ref{fig:system_pipeline}.
This work utilizes \acp{GP} as the data-driven models of vehicle dynamics due to their many advantages \citep{jain_bayesrace_2020,rodriguez_learning_2021,hewing2018cautious}.



\section{Ensemble Gaussian Process}
\label{section:EGP}



As the model changes as a function of a hidden parameter $\theta^n$, estimating a single \ac{GP} for all operating conditions may not be possible -- see simulation results in Section~\ref{sec:results}. Thus, we propose to use an ensemble of \acp{GP}. Given $N$ \acp{GP} $\mathcal{GP}^1 \dots \mathcal{GP}^N$, we would like to obtain a model which is an ensemble of these \acp{GP}, denoted by $\mathcal{GP}^E$. The resulting model is a valid \ac{GP} as it is a linear combination of $N$ \acp{GP}. Given a vector of weights $w = [w_1, w_2, \dots, w_N]^T$, we can find the maximum likelihood estimate of output given the input $\hat{y}_{k+1|x_k}$ and its variance $\hat{\sigma}^2_{k+1|x_k}$ as follows:
\begin{subequations}\label{eq:ensemble}
\begin{align}
\hat{y}_{k+1|x_k} &= \sum_{n=1}^{N} w_n\hat{\mu}_{k+1|x_k}^n\\
\hat{\sigma}^2_{k+1|x_t} &= \sum_{n=1}^{N} w_n^2(\hat{\sigma}_{k+1|x_k}^n)^2
\end{align}
\end{subequations}
Where $\hat{\mu}_{k+1|x_k}^n$ is the mean of the $n^{th}$ \ac{GP}
$\mathcal{GP}^n$ and $\hat{\sigma}_{k+1|x_k}^n$ is the corresponding standard deviation of that \ac{GP}.

Since differentiation is a linear operator, we can approximate the ensembled mean and variance by using Taylor expansion about a nominal input point $x^l$. Let $\nabla_{x^i}f_{\mu}^n$ be the Jacobian of the mean function $f_\mu^n$ of the $n^{th}$ \ac{GP} with inputs $x^i$. We can use this jacobian to find a linear approximation of $\hat{\mu}^n$ as $\tilde{\mu}^n \approx x^l + \Delta_x^T\nabla_xf_{\mu}^n$.

As this is now a linear function of $x$, we can now express $\hat{y}_{k+1|x_k}$ as $\tilde{y}_{k+1|x_k}$, the linear combination of the $\tilde{\mu}^n$ functions as follows:
\begin{equation}
    \tilde{y}_{k+1|x_k} = x^l + \Delta_x^T\sum_{n=1}^{N} w_n\nabla_xf_{\mu}^n
    \label{equation:EGPprediction}
\end{equation}

\subsection{Model weights adaptation}
\vspace{-6pt}
\label{sec:WeightsEstimation}
To calculate the prediction of the ensembled model, we need to first calculate the vector of weights 
\begin{align}
    w=[w_1, w_2, \dots, w_N]^T.    
\end{align}
Let $\hat{y}^n$ be the prediction of the $n^{th}$ \ac{GP} $\hat{y}^n = f_\mu^n(x^i)$, where $x^i$ is the input to the \ac{GP}. Let $y$ be the true value of the output. Given the history of the length $K$ of the output-input pairs $[(y_{k-1}, x^i_{k-1}), \dots, (y_{k-K}, x^i_{k-K})]$ our goal is to calculate the combination of models that provide the best representation over the history. We can formulate this problem as the following optimization program:
\begin{subequations}
    \label{eq:WeightsEstimation}
    \begin{align}
        w^* = \argminB_{w} \quad & {\|Y - Fw\|}_2^2 + \alpha{\|w - w_{k-1}\|}_1 \\
        \textrm{subject to} \quad & 0 \leq w \leq 1, \\
        & \textbf{1}w^T = 1,
    \end{align}
\end{subequations}

where $Y$ is the vector of true output values
\begin{align}
    Y =
    \begin{bmatrix}
        y_{t-1} &
        y_{t-2} &
        \cdots&
        y_{t-K}
    \end{bmatrix}^T,
\end{align}
$F$ is a matrix of predictions from all of the models over the whole history
\begin{align}
F =
\begin{bmatrix}
f_\mu^1(x^i_{k-1}) & \cdots & f_\mu^N(x^i_{k-1})\\
f_\mu^1(x^i_{k-2}) & \cdots & f_\mu^N(x^i_{k-2})\\
\vdots &  & \vdots\\
f_\mu^1(x^i_{k-K}) & \cdots & f_\mu^N(x^i_{k-K})\\
\end{bmatrix},
\end{align}
and $\alpha$ is a regularization parameter that minimizes the distance between the previous estimate of the weights $w_{k-1}$ and the new weights. 


\section{Ensemble Gaussian Process Model Predictive Control (EGP-MPC)}\label{sec:MPC}
\subsection{System modeling}
\vspace{-6pt}
\label{sec:sys_models}

In this section, we present the system identification strategy. We consider a vehicle with states:
\begin{align}
    x = [p^x, p^y, v_x, \psi, v_y, \omega, \delta],
\end{align} 
where $p^x$ and $p^y$ are the position in Cartesian coordinates, $\psi$ is the orientation, $v_x$ and $v_y$ are the longitudinal and lateral velocities, $\omega$ is the yaw rate, and $\delta$ is the steering angle. The control input is $u=[F^x, \Dot{\delta}]$, where $F^x$ is the engine drive force, and $\dot{\delta}$ is the steering velocity. 
To estimate a discrete-time model, we exploit the kinematic equations of motion and construct a data-driven model of the dynamics using \acp{GP}. The main advantage of using \acp{GP} is that it is possible to reason about the uncertainty of the model prediction. 

The kinematic equations of motion are defined as follows:
\begin{subequations}
\label{eq:kin}
\begin{align}
    \Dot{p}_x &= v_x\cos(\psi) - v_y\sin(\psi),
    \label{equation:kin_1}\\
    \Dot{p}_y &= v_x\sin(\psi) + v_y\cos(\psi),
    \label{equation:kin_2}\\
    \Dot{\psi} &= \omega,
    \label{equation:kin_3}\\
    \delta &= \Dot{\delta}.
    \label{equation:kin_4}
\end{align}
\end{subequations}
To describe the dynamics of the system using equations of motion, we would need to perform a system identification campaign for all of the physical parameters. However, system identification for tire parameters is time-consuming as it requires designing specific experiments for data collection as demonstrated by \cite{van2018vehicle}.
Therefore, we choose to use a data-driven approach using \acp{GP}. We discretize \eqref{eq:kin} and model each of the dynamic states $v_x$, $v_y$, and $\omega$ as an independent \ac{GP} directly in a discretized form. Discretized model equations have the following form
\begin{equation}\label{eq:sys_model_all}
\begin{aligned}
    p_x[k+1] &= f_{px} = p_x[k] + (v_x\cos(\psi) - v_y\sin(\psi))dt,\\
    p_y[k+1] &= f_{py} = p_y[k] + (v_x\sin(\psi) + v_y\cos(\psi))dt,\\
    v_x[k+1] &= f_{vx} = v_x[k] + f_{\mu,vx}(v_x, v_y, \omega, \delta, F^x, \Dot{\delta}; \theta),\\
    \psi[k+1] &= f_{\psi} = \psi[k] + \omega dt,\\
    v_y[k+1] &= f_{vy} = v_y[k] + f_{\mu,vy}(v_x, v_y, \omega, \delta, F^x, \Dot{\delta}; \theta),\\
    \omega[k+1] &= f_{\omega} = \omega[k] + f_{\mu,\omega}(v_x, v_y, \omega, \delta, F^x, \Dot{\delta}; \theta),\\
    \delta[k+1] &= f_{\delta} = \delta[k] + \Dot{\delta}dt,
\end{aligned}
\end{equation}
where $dt$ is the discretization time step, and $f_{\mu,vx}$, $f_{\mu,vy}$, and $f_{\mu,\omega}$ are mean functions of the \acp{GP} for the longitudinal velocity, lateral velocity and yaw-rate respectively. In the above equation, $\theta$ represents the friction that affects the dynamics of the system. We can write the system model from \eqref{eq:sys_model_all} in a more compact way as 
\begin{align}\label{eq:sys_model}
    x_{k+1} = f(x_{k}, u_{k}; \theta) = 
    \begin{bmatrix}
        f_{px}(x_{k}, u_{k})\\
        f_{py}(x_{k}, u_{k})\\
        f_{vx}(x_{k}, u_{k}; \theta)\\
        f_{\psi}(x_{k}, u_{k}; \theta)\\
        f_{vy}(x_{k}, u_{k}; \theta)\\
        f_{\omega}(x_{k}, u_{k})\\
        f_{\delta}(x_{k}, u_{k})
    \end{bmatrix}.
\end{align}

As we discussed in the Section~\ref{sec:problem_formulation} we propose an ensemble of models to solve the problem of driving under changing road friction $\theta$. We assume that we have $N$ datasets defined as in~\eqref{eq:data}, each collected under different friction parameters $\theta^1, \ldots, \theta^N$. Leveraging these datasets we compute a library of models $f^1(x, u; \theta^1), \dots, f^N(x, u; \theta^N)$ using~\eqref{eq:sys_model}. Then, we compute model weights $w$ using \eqref{eq:WeightsEstimation} and the resulting ensemble model is:
\begin{align}\label{eq:ensemble_model}
    \hat{f}(x, u; \theta) = \sum^N_{n=1}w_n f(x, u; \theta^n).
\end{align}
%
%
%
%
%
\subsection{Control Synthesis}
\vspace{-6pt}
In this section, we describe the EGP-MPC algorithm. To reduce the computational complexity we leverage a linearized version of the ensemble model from \eqref{eq:ensemble_model}. The linearization of the ensemble of \acp{GP} from Section~\ref{section:EGP} can be used to construct a linearized system model about the nominal point $x^l$. This is achieved by computing the system Jacobian using the values of linearized dynamic states from \eqref{equation:EGPprediction}. Thus at every time $k$, we choose an operating trajectory $x^l_{t|k}$, $u^l_{t|k}$, $t \in \{0,1,\dots,T\}$, where $T$ is the prediction horizon, around which we linearize the system defined by the models $f^1(x, u; \theta^1), \dots, f^N(x, u; \theta^N)$. Then, we 
compute model matrices
\begin{align}
    [A^1_{t|k}, B^1_{t|k}, C^1_{t|k}; \dots; A^N_{t|k}, B^N_{t|k}, C^N_{t|k}],    
\end{align}
where $N$ is the number of models, and
\begin{equation}
    \label{eq:ABC}
    \centering
    \begin{aligned}
        A^n_{t|k} &=  \nabla_xf^n(x^l_{t|k}, u^l_{t|k}; \theta^n)\\
        B^n_{t|k} &=  \nabla_uf^n(x^l_{t|k}, u^l_{t|k}; \theta^n),\\
        C^n_{t|k} &=  f^n(x^l_{t|k}, u^l_{t|k}; \theta^n) - A^n_{t|k}x^l_{t|k} - B^n_{t|k}u^l_{t|k}.
    \end{aligned}
\end{equation}
Finally, using weights $w$ we create a linearized ensemble model in the form: 
\begin{align}\label{equation:ensemble_model_lin}
    x_{t+1|k} = \hat{A}_{t|k}x_{t|k} + \hat{B}_{t|k}u_{t|k} + \hat{C}_{t|k}, 
\end{align}
where
\begin{equation}
    \label{eq:ATVE}
    \begin{aligned}
        \hat{A}_{t|k} = \sum^N_{n=1} w_n A^n_{t|k}, &\quad
        \hat{B}_{t|k} = \sum^N_{n=1} w_n B^n_{t|k}, \\
        \hat{C}_{t|k} = &\sum^N_{n=1} w_n C^n_{t|k}.
    \end{aligned}
\end{equation}

Next, we leverage the above linearized ensemble model matrices $\hat{A}_{t|k}$, $\hat{B}_{t|k}$, $\hat{C}_{t|k}$ to design the EGP-MPC. At every time $k$, given the initial state $x_k$ we solve the following finite-time optimal control problem (FTOCP):
\begin{subequations}
\label{eq:EGP_MPC}
\begin{align}
    \textbf{u}^*, \textbf{x}^* = \quad \quad \quad  & \notag \\ 
    \argminB_{u_{t|k}, x_{t|k}} \quad & \sum_{t=1}^{T-1}(||x_{t|k} - x^r_{t|k}||_{Q}+||u_{t|k}||_R) \notag \\
    + & \sum_{t=2}^{T-1}||u_{t|k} - u_{t+1|k}||_{R_d} +||x_{T|k}-x^r_{T|k}||_{Q_T} \\
    \textrm{subject to} \quad &x_{0|k} = x_{k},\\
    &x_{t+1|k} = \hat{A}_{t|k}x_{t|k} + \hat{B}_{t|k}u_{t|k} + \hat{C}_{t|k},\\
    &x_{t|k} \in \mathcal{X}\\
    &u_{t|k} \in \mathcal{U},\\
    &\forall t \in \{0, 1, \cdots, T-1\},
\end{align}
\end{subequations}
where $\textbf{x}^* = [x_{t|k}^*, \dots, x_{t+T|k}^*]$, $\textbf{u}^* = [u_{t|k}^*, \dots, u_{t+T|k}^*]$, and $\textbf{x}^r = [x_{t|k}^r, \dots, x_{t+T|k}^r]$ is a reference trajectory which in our case consists of reference position, velocity, and orientation, $\mathcal{X}$ is the state space, $\mathcal{U}$ is the input space, and the norm $\|z\|_Q:=z^TQz$. The proposed approach is shown in Algorithm~\ref{alg:egp_mpc}. First, we initialize the GP library $GP_{lib}$ and all necessary variables (lines~1--4). At each time step, we compute the matrices for the ensemble using~\eqref{eq:ATVE} (lines~6--9). Then, we solve the FTOCP~\eqref{eq:EGP_MPC} and store the optimal solution (lines~10--11). 
Next we store the current vector of inputs to the \acp{GP} $x^i_b$ and apply to the system the first element of the optimizer vector $\textbf{u}^*$ from (20) (lines~12--13). We then observe the system state and store the vector of inputs to the \acp{GP} after applying the control input $x^i_a$ (line~14). Now, we calculate the difference between $x^i_a$ and $x^i_b$ and store the first three elements which are the dynamic state transition (lines~15--16). Finally, we update the history of measurements $H$, calculate the weights $w$ using a sliding window over $H$ as in \eqref{eq:WeightsEstimation}, and store $w$ as $w_{prev}$ for the next iteration (lines~17--19).

\begin{algorithm}[hb]
\begin{algorithmic}[1]
 \caption{EGP-MPC}
 \label{alg:egp_mpc}
 \State ${GP}_{lib} \gets \mathcal{GP}^1 \dots \mathcal{GP}^N$
 \State $x_{prev} \gets$ $[0, 0, \dots, 0]$, $u_{prev} \gets$ $[0, 0, \dots, 0]$
 \State $w \gets$ $[\frac{1}{N}, \frac{1}{N}, \dots, \frac{1}{N}]$, $w_{prev} \gets$ $[\frac{1}{N}, \frac{1}{N}, \dots, \frac{1}{N}]$
 \State $H \gets$ $\emptyset$
 \For{each time step $k$}
    \For{each $n^{th}$ GP}
        \State Update $A^n_{t|k}$, $B^n_{t|k}$ and $C^n_{t|k}$ using \eqref{eq:ABC}
    \EndFor
    \State Update $\hat{A}_{t|k}$, $\hat{B}_{t|k}$, $\hat{C}_{t|k}$ using \eqref{eq:ATVE} and $w$
    \State $u^*$, $x^*$ $\gets$ solution of \eqref{eq:EGP_MPC}
    \State $x_{prev}$, $u_{prev}$ $\gets$ $x^*$, $u^*$
    \State $x^i_a \gets [v^x, v^y, \omega, \delta, F^x, \Dot{\delta}]$
    \State Apply $u^*$ to the vehicle for a $dt$
    \State $x^i_b \gets [v^x, v^y, \omega, \delta, F^x, \Dot{\delta}]$
    \State $d \gets (x^i_b - x^i_a)$
    \State $y \gets [d[0], d[1], d[2]]$
    \State Append tuple $(y, x^i_1)$ to $H$
    \State Update $w$ using $w_{prev}$ and $H$ according to \eqref{eq:WeightsEstimation}
    \State $w_{prev} \gets w$
 \EndFor
\end{algorithmic}
\end{algorithm}

\section{Results}
\label{sec:results}

We test our approach on the F1Tenth gym simulator \cite{okelly_f1tenth_2020} with a multi-body model from \cite{althoff_commonroad_2017} and vehicle parameters of the vehicle ID: 1 from \cite{althoff_commonroad_2017}. With this setup, we use a full-scale car simulator with complex multi-body dynamics to test the performance of our approach which is using a simpler model for the controller. 

 We then show that using one \ac{GP} is not sufficient for driving on different conditions, as the ensemble of \acp{GP} can achieve lower  prediction errors in the environment that they are trained on. We also show that the ensemble \ac{GP} can achieve similar prediction error for new, unseen surface conditions. We validated our controller by driving on a custom racetrack with a time-optimal raceline optimized using \cite{christ2021time}. Additionally, we tested our controller by tracking the Sao Paolo raceline from \cite{Betz2022_RacingSurvey} as well as performing several dual-lane change \cite{iso2011passenger} standard maneuvers. Finally we show that the learned controller can generalize to new tracks and surface conditions that it was not trained on.

All code for the EGP-MPC was implemented in Python. GPyTorch \cite{gardner_gpytorch_2021} was used for \ac{GP} training and inference. GPyTorch provides fast variance prediction via LanczOs Variance Estimates (LOVE) \cite{noauthor_180306058_nodate} as well as fast kernel operations through KeOps \cite{ragan-kelley_halide_2017}. GPyTorch also interfaces with Pytorch autograd functionality allowing for easy computation of Jacobians, gradients, and Hessians. The FTOCP, as well as the model weight estimation problem, was solved using OSQP \cite{stellato_osqp_2020} with CVXPY \cite{diamond2016cvxpy, agrawal2018rewriting} as the interface. CVXPY provides an easy-to-use interface to OSQP and is capable of generating C code for optimization problems using CVXPYgen. Using the linearized models of 2 \acp{GP} and a sliding window size $K = 11$, the EGP-MPC achieves an average control frequency of 30Hz. All code can be found here: \texttt{https://github.com/atomyks/multisurface-racing}.

\subsection{Comparison with Standard GP}
\vspace{-6pt}
\label{sec:comp_with_standard_GP}

In this section, we show that the ensemble \ac{GP} outperforms a single \ac{GP} when the vehicle is operating on surfaces with different friction. We also show that an ensemble \ac{GP} model is able to generalize to surfaces with frictions it was not trained on.

We start by creating five datasets [$D^A$, $D^B$, $D^T$, $D^C$, $D^D$], for five friction values [0.3, 0.6, 0.7, 0.8, 1.0] respectively. First, we define bounded subsets of the state and input spaces. Then, we randomly sample states and inputs from these subsets. We drive the car initialized at the sampled state for 0.1 seconds, while applying the sampled control input. After each time step ($k = 0.02$ seconds), we store a sequence of two consecutive inputs and outputs for the \acp{GP}, i.e., we store tuples $[(y_{k-1}, x^i_{k-1}), (y_{k}, x^i_{k})]$ where
\begin{align*}
    &x^i_{k} = [v_x[k], v_y[k], \omega[k], \delta[k], F^x[k], \Dot{\delta}[k]],\\
    &y_{k} = [v_x[k] - v_x[k-1], v_y[k] - v_y[k-1], \omega[k] - \omega[k-1]].
\end{align*}
We repeat this process until we have collected 1000 tuples. Then, we split the datasets [$D^A$, $D^B$, $D^C$, $D^D$] into training sets ($70\%$ of samples) [$D_{train}^A$, $D_{train}^B$, $D_{train}^C$, $D_{train}^D$], and validation sets ($30\%$ of samples) [$D_{val}^A$, $D_{val}^B$, $D_{val}^C$, $D_{val}^D$], while $D^T$ is kept as a testing set.

\begin{figure}[tb]
    \centering
    \vspace{-5pt}
    \includegraphics[trim={0.0cm 0.0cm 0.0cm 0.0cm},clip,width=8.4cm]{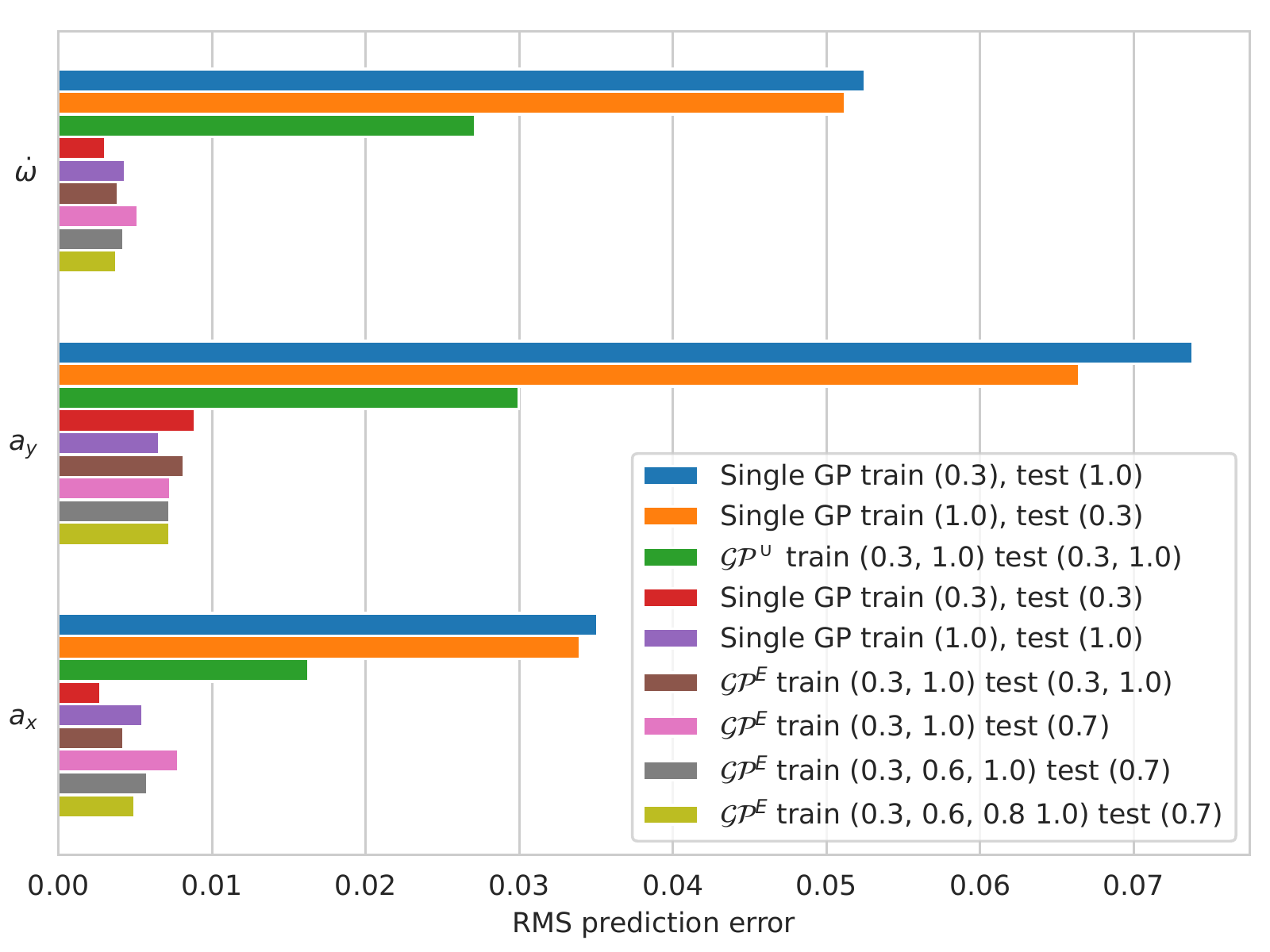}
    \vspace{-10pt}
    \caption{Comparison of root-mean-square (RMS) error for single \acp{GP} trained and validated on datasets with different frictions (Blue, Orange, Red, Purple), single \ac{GP} trained on a mixture of frictions and validated on that mixture (Green), our proposed approach which ensembles multiple single \acp{GP} trained on different frictions and validated on a mixture of those frictions (Brown), and our proposed approach validated on unseen frictions with multiple \acp{GP} models (Pink, Grey, Olive).}
    \label{fig:Prediction_error}
\vspace{-2pt}
\end{figure}

We train five \ac{GP} models [$\mathcal{GP}^A$, $\mathcal{GP}^B$, $\mathcal{GP}^C$, $\mathcal{GP}^D$, $\mathcal{GP}^\cup$] trained on the datasets [$D_{train}^A$, $D_{train}^B$, $D_{train}^C$, $D_{train}^D$, $D_{train}^A \cup D_{train}^D$]. To validate our trained models, we test our \acp{GP} on their respective validation sets [$D_{val}^A$, $D_{val}^B$, $D_{val}^C$, $D_{val}^D$, $D_{val}^A \cup D_{val}^D$]. In Figure~\ref{fig:Prediction_error}, we only visualize $\mathcal{GP}^A$, $\mathcal{GP}^D$ and $\mathcal{GP}^\cup$ for clarity. From Figure~\ref{fig:Prediction_error}, we can see that $\mathcal{GP}^A$, $\mathcal{GP}^D$ perform well on their respective validation sets but perform poorly when validated on datasets collected from a friction different from the friction encountered during training. We can also see that $\mathcal{GP}^\cup$ which was trained on data collected from both friction sets $D_{train}^A \cup D_{train}^D$ performs better than the worst-case performance of $\mathcal{GP}^A$ and $\mathcal{GP}^D$, but much worse than the best-case performance of $\mathcal{GP}^A$ and $\mathcal{GP}^D$. 

Our proposed approach using $\mathcal{GP}^E$ which was ensembled from  $\mathcal{GP}^A$ and $\mathcal{GP}^D$ outperforms $\mathcal{GP}^\cup$. To validate the ensemble approach, we first compute weights $w$ according to \eqref{eq:WeightsEstimation} using models $\mathcal{GP}^{A}$ and $\mathcal{GP}^{D}$, and the $(y_{k-1}, x^i_{k-1})$ samples from $D^A_{val} \cup D^D_{val}$. Then using weights $w$ and the models $\mathcal{GP}^{A}$ and $\mathcal{GP}^{D}$ we calculate the ensembled prediction using $x^i_{k}$ and then validate against $y_{k}$ from $D^A_{val} \cup D^D_{val}$. As we can see, the ensembled model $\mathcal{GP}^E$ performs much better than the worst-case performance of each of the \acp{GP}, $\mathcal{GP}^A$ and $\mathcal{GP}^D$, and is almost as good as their best-case performance.

To test generalization to unseen friction parameters, we use dataset $D^T$ with a coeficient of the friction 0.7. Then, we test the model ensembled from $\mathcal{GP}^A$ and $\mathcal{GP}^D$ on the dataset $D^T$. Note that such a friction value is between the friction values from dataset $D^A$ and $D^D$. As we use a convex combination of models -- see Section~\ref{sec:WeightsEstimation} -- our method can only predict the evolution of the vehicle on unseen frictions that are between the frictions used to construct the training datasets. It can be seen that the accuracy of the prediction does not change much when we introduce a new friction surface. 

To test if the accuracy of ensembled model $\mathcal{GP}^E$ on the unseen friction surface increases as we increase the number of ensembled \acp{GP}, we try three variants of $\mathcal{GP}^E$: $\mathcal{GP}^E$ ensembled from $[\mathcal{GP}^A, \mathcal{GP}^D$], $\mathcal{GP}^E$ ensembled from $[\mathcal{GP}^A, \mathcal{GP}^B, \mathcal{GP}^D$], and $\mathcal{GP}^E$ ensembled from $[\mathcal{GP}^A, \mathcal{GP}^B, \mathcal{GP}^C, \mathcal{GP}^D$]. We can see that the accuracy of the ensembled model increases and tends towards the training accuracy as we ensemble more \acp{GP}. This happens because the newly added \acp{GP} cover more of the parameter space and are closer to the unseen friction. 

\subsection{Multi-surface Driving and Generalization}\label{subsec:experiment_var_speed}
\begin{figure}[tb]
    \centering
    \includegraphics[trim={0.0cm 0.0cm 0.5cm 0.0cm},clip,width=8.4cm]{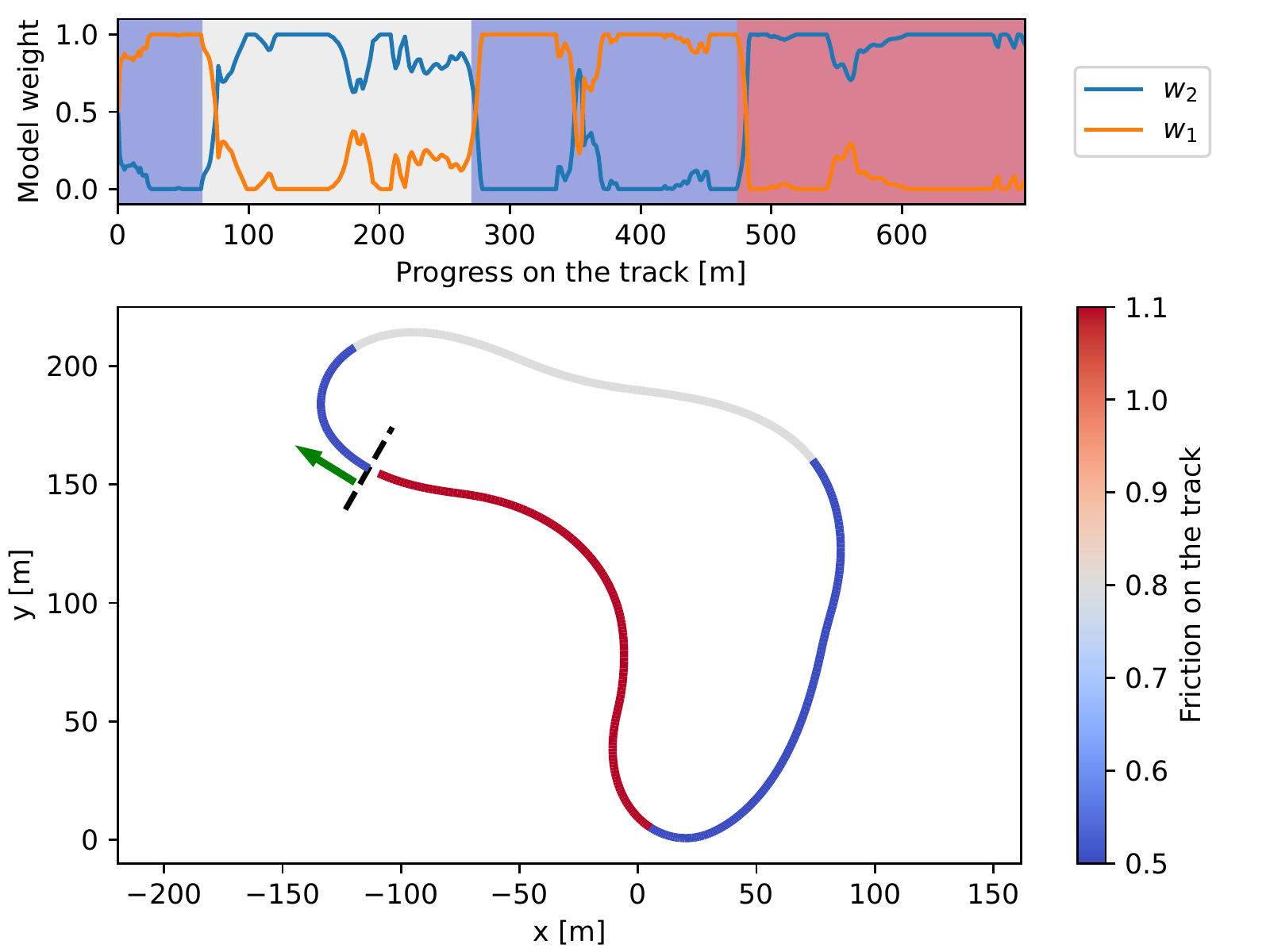}
    \vspace{-10pt}
    \caption{Reference trajectory with color-coded friction regions (Bottom) and model weights $w_1$ (Low-friction model) and $w_2$ (High-friction model) used by the EGP-MPC (Top). Note that the EGP-MPC is mostly using the high-friction model driving on high-friction surfaces, and the low-friction model driving on the lower friction surface. }
    \label{fig:vel_profile-reference}
    \vspace{-8pt}
\end{figure}

\begin{figure}[tb]
    \centering
    \includegraphics[trim={0.5cm 0.0cm 1.5cm 0.0cm},clip,width=8.4cm]{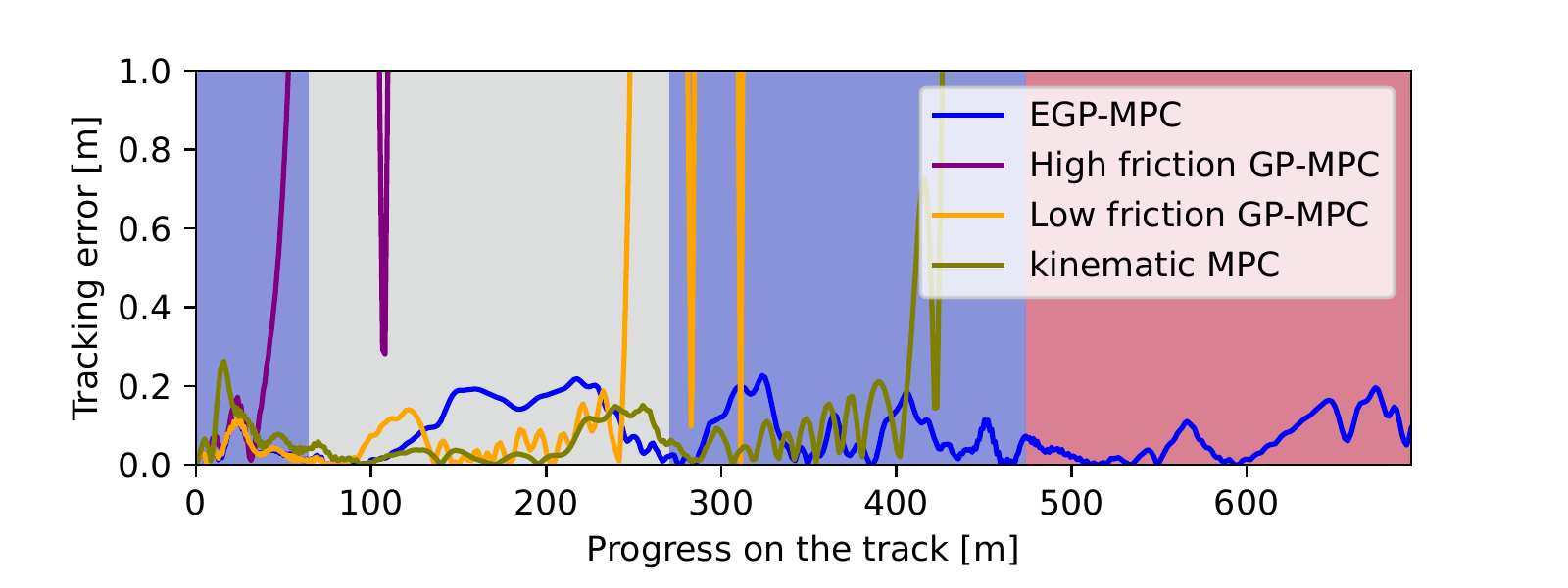}
    \vspace{-10pt}
    \caption{Tracking error associated with the proposed strategy and baselines. Note that all three baselines fail to complete the lap. The GP-MPC trained on high-friction fails on the low-friction surface after 40m; GP MPC trained on low-friction fails after 240m on higher friction surface, and kinematic MPC fails on the low-friction surface after 400m. On the other hand, the proposed EGP-MPC is able to complete the lap.}
    \label{fig:vel_profile-tracking_error}
    \vspace{-9pt}
\end{figure}

\begin{figure}[tb]
    \centering
    \includegraphics[trim={0.8cm 0.0cm 0.1cm 0.0cm},clip,width=8.4cm]{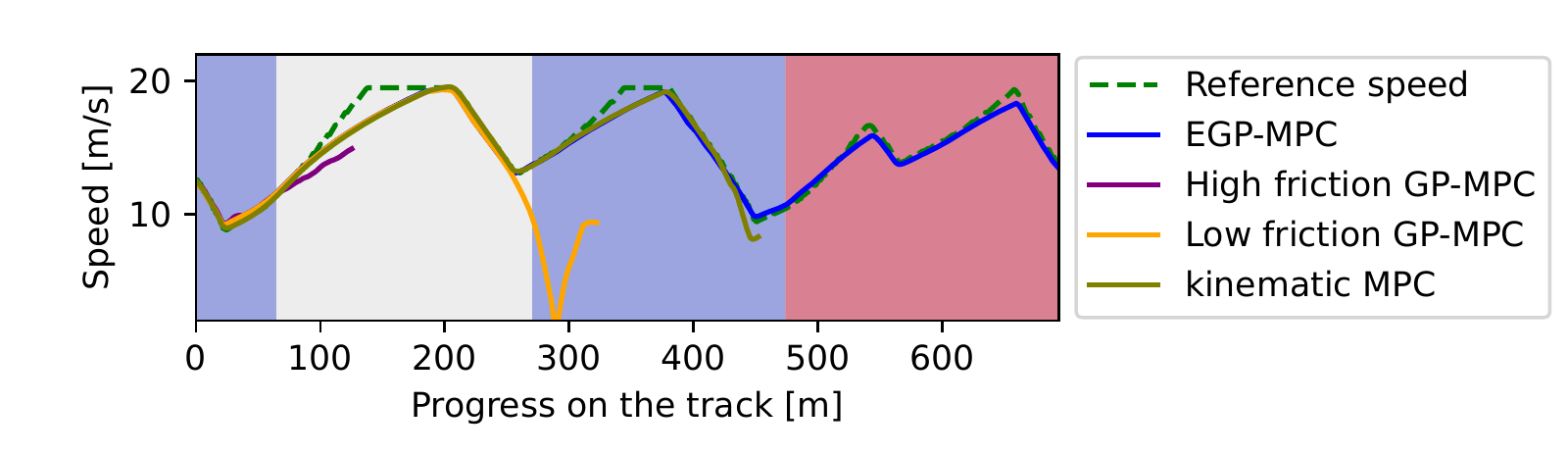}
    \vspace{-10pt}
    \caption{Reference speed and velocity of the closed-loop system for different controllers. Notice that only the proposed strategy is able to track the speed profile.}
    \label{fig:vel_profile-speed}
\end{figure}

\vspace{-6pt}
We test the proposed strategy on three scenarios:

\textbf{Driving on a track with varying speed:} In this experiment, we drive the car following an optimized trajectory on a custom racetrack. We trained two models for frictions $\theta^1 = 0.5$ and $\theta^2 = 1.1$. The training data are collected by driving on the track with single surface friction and gradually increasing speed. At test time, there are three friction zones ($0.5, 0.8, 1.1$) and the proposed strategy computes the weight $w_1$ and $w_2$ using~\eqref{eq:WeightsEstimation}. The friction zones on the track and the weights used by EGP-MPC are shown in Figure~\ref{fig:vel_profile-reference}. We can see that our controller is correctly choosing the GP models for the specific friction as the weights used in the ensemble correspond to the different friction surfaces. We compared the EGP-MPC with single GP-MPC and kinematic MPC controllers. Driving speeds and tracking errors are presented in Figure~\ref{fig:vel_profile-speed} and \ref{fig:vel_profile-tracking_error}. The EGP-MPC is the only controller that successfully follows the trajectory while tracking the desired speed.

\begin{figure}[ht]
    \centering
    \includegraphics[trim={0.0cm 0.0cm 0.5cm 0.0cm},clip,width=8.4cm]{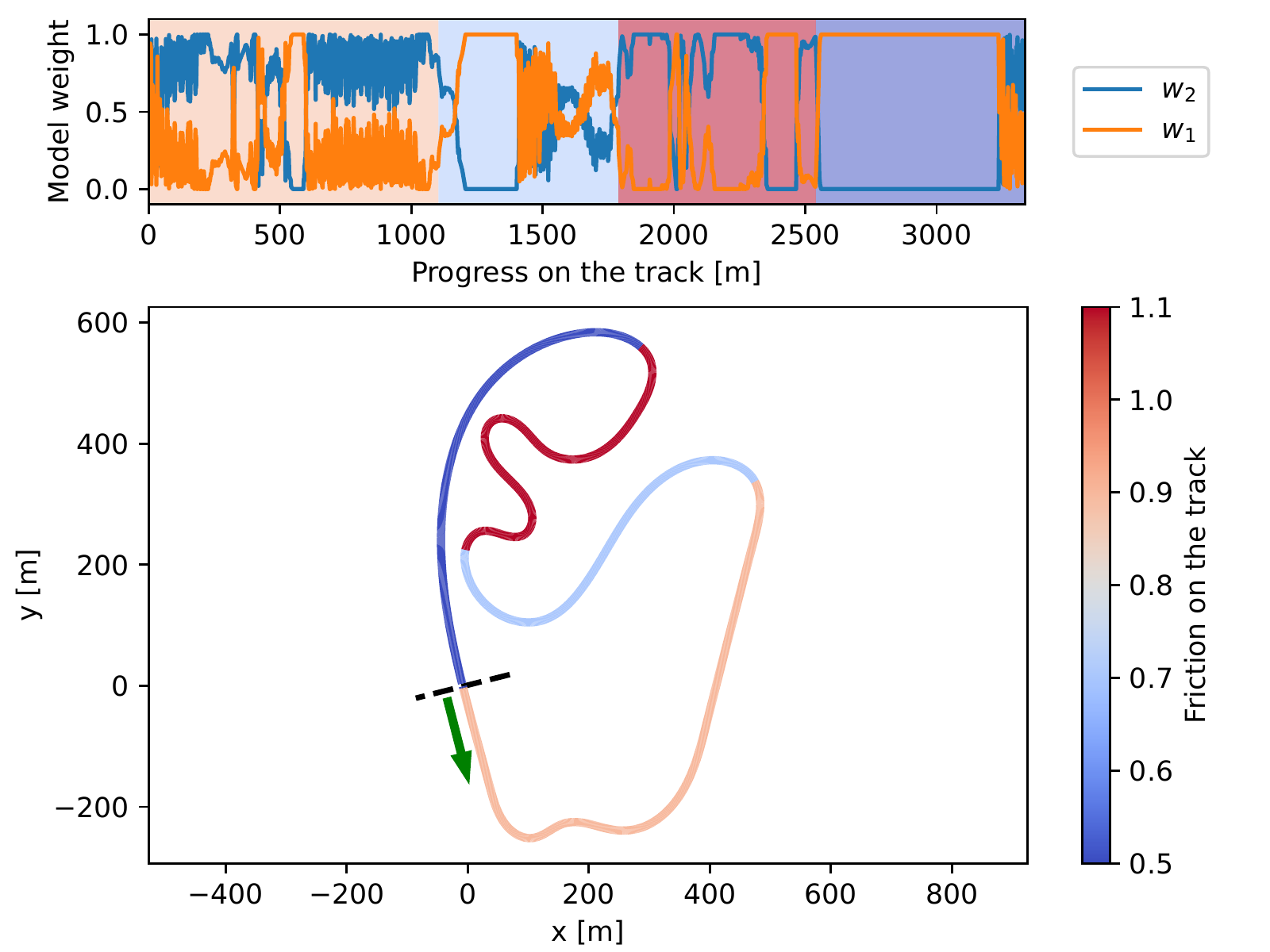}
    \vspace{-10pt}
    \caption{Reference trajectory with color-coded friction regions (Bottom) and model weight $w_1$ (Low-friction model) and $w_2$ (High-friction model) used by the EGP-MPC (Top). 
    }
    \label{fig:generalization-reference}
\vspace{-2pt}
\end{figure}

\begin{figure}[ht]
    \centering
    \includegraphics[trim={0.5cm 0.0cm 1.5cm 0.0cm},clip,width=8.4cm]{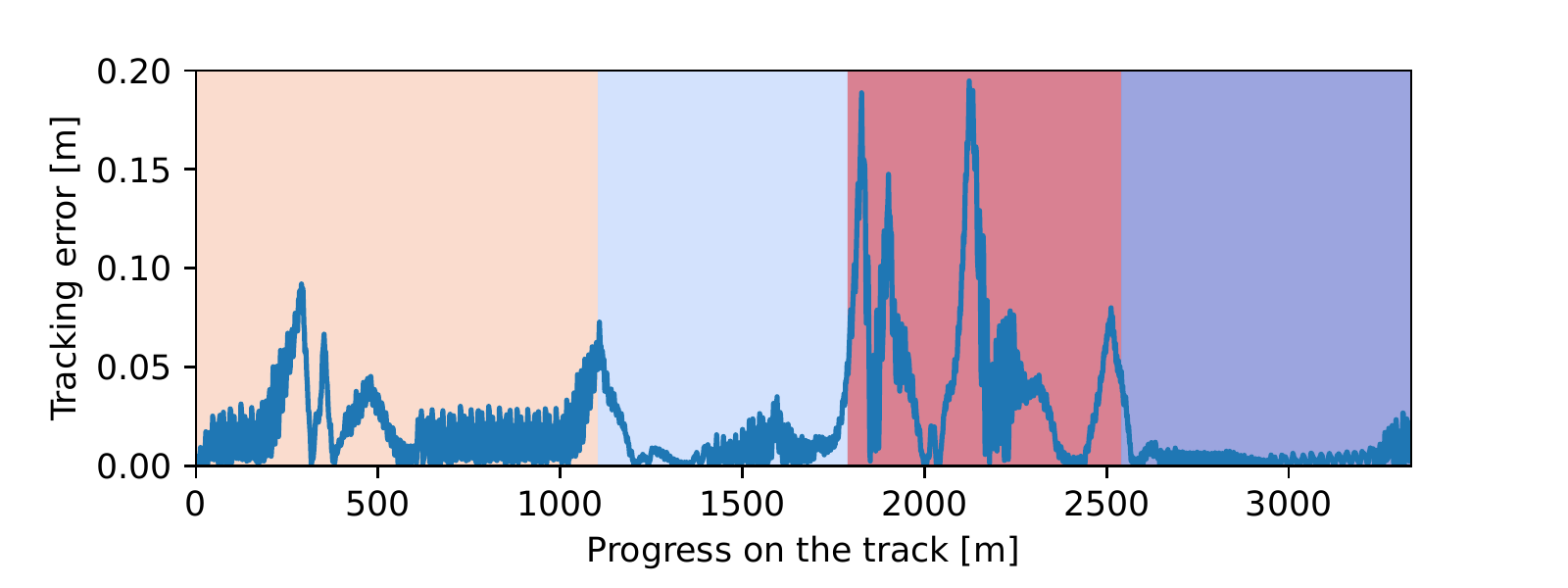}
    \caption{Tracking error while driving on the track that the model was not trained on.}
    \label{fig:generalization-tracking_error}
\end{figure}

\textbf{Testing on an unseen track:} To show that the controller can generalize across multiple trajectories, we test the EGP-MPC that we trained in the previous experiment on the Sao Paolo ractrack shown in Figure~\ref{fig:generalization-reference}. We do not retrain the models on any parts of the new track. The reference speed for this experiment is constant at $14 $ m/s. As we can see in Figures~\ref{fig:generalization-reference} and \ref{fig:generalization-tracking_error}, the algorithm is still able to perform well even when driving on a completely new track.

\begin{figure}[tb]
    \centering
    \includegraphics[trim={0.0cm 0.0cm 0.5cm 0.0cm},clip,width=8.4cm]{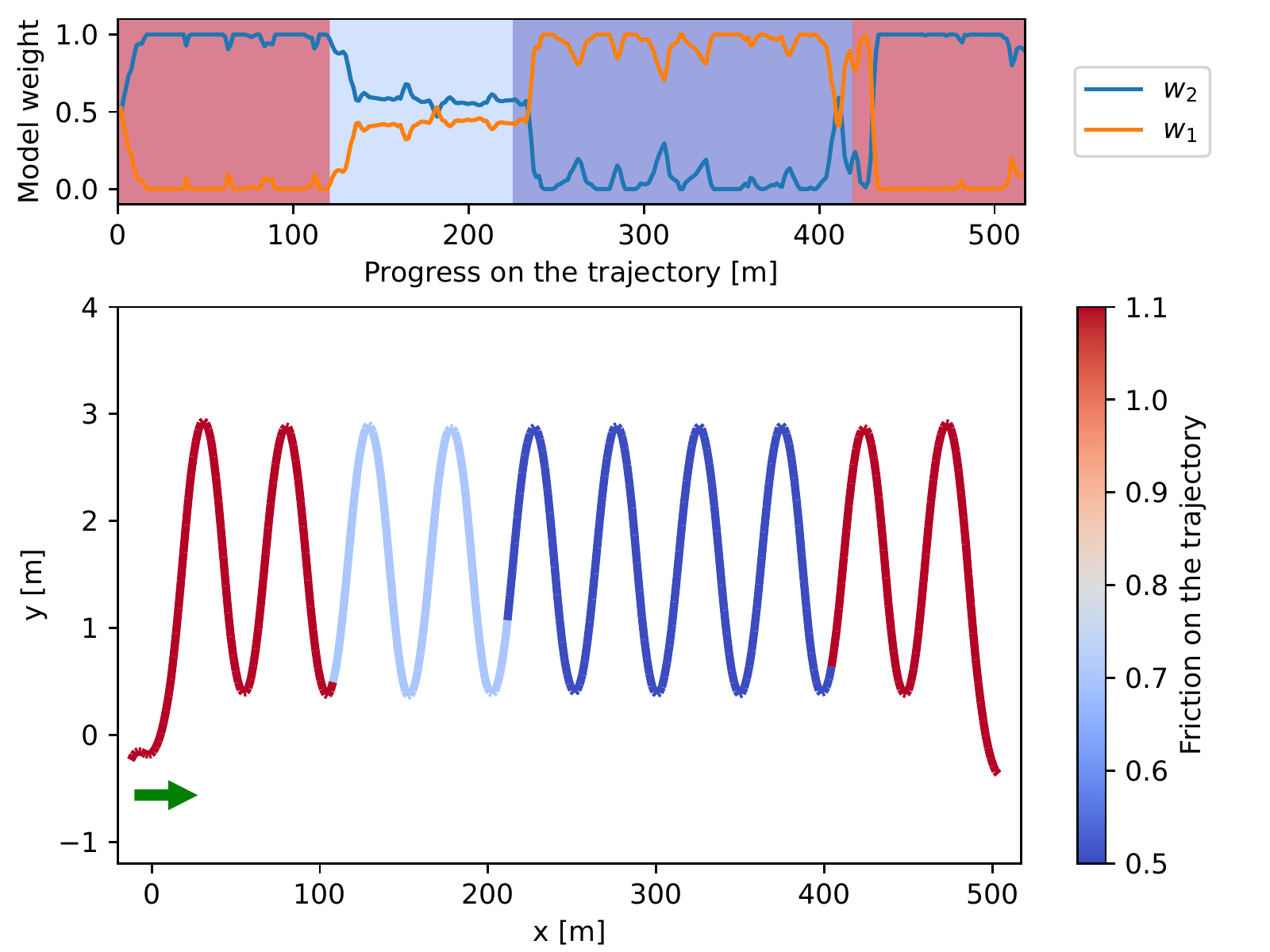}
    \vspace{-10pt}
    \caption{Reference trajectory with color-coded friction regions (Bottom) and model weights $w_1$ (Low-friction model) and $w_2$ (High-friction model) used by the EGP-MPC (Top). Note that the proposed EGP-MPC correctly chooses the GP models trained for the corresponding frictions, and it uses a combination of the two models when the friction is in between.}
    \label{fig:LineChange-reference}
\vspace{-8pt}
\end{figure}

\begin{figure}[tb]
    \centering
    \includegraphics[trim={0.5cm 0.0cm 1.5cm 0.0cm},clip,width=8.4cm]{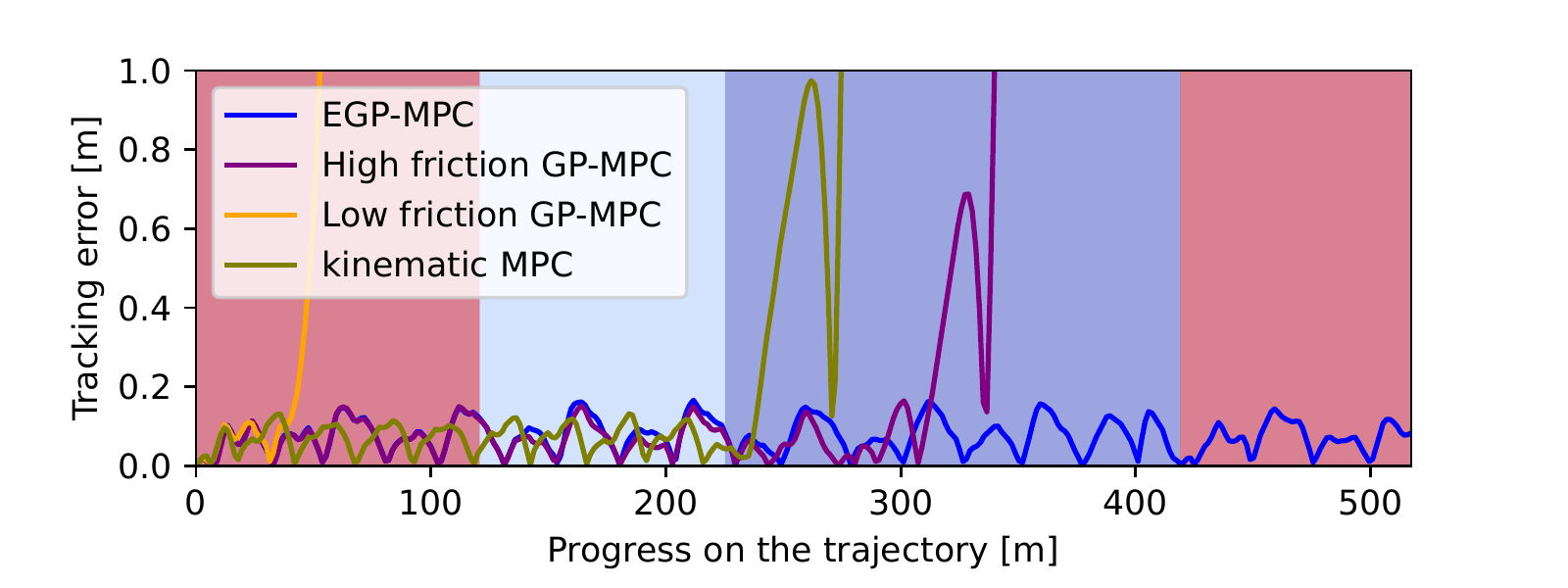}
    \vspace{-10pt}
    \caption{Tracking error associated with the proposed strategy and baselines.}
    \label{fig:LineChange-trajectories}
\vspace{-3pt}
\end{figure}

\textbf{Lane change maneuver:} We also test our algorithm on the standard dual lane change maneuver \cite{iso2011passenger}. For this experiment, we train two new models by driving this maneuver at different friction coefficients: the first model is trained on data collected at a coefficient of friction $\theta^1 = 0.5$, while the second model is trained on data collected at a coefficient of friction $\theta^2 = 1.1$. For testing, we split up the maneuver into four segments of friction coefficients $[1.1, 0.7, 0.5, 1.1]$. The reference speed is set to $15 m/s$ and the proposed strategy computes the weight $w_1$ and $w_2$ using~\eqref{eq:WeightsEstimation}. Figure~\ref{fig:LineChange-reference} shows the reference trajectory, friction coefficients of the trajectory, and the weight estimation by EGP-MPC. During the lane change maneuver, the dynamics of the vehicle are well excited. Therefore, the weight in the ensemble corresponds to the friction with less fluctuation than in Figure~\ref{fig:generalization-tracking_error}. A comparison of the tracking performance of different algorithms can be seen in Figure~\ref{fig:LineChange-trajectories}. It is important to underline that all tested algorithms except for the proposed EGP-MPC failed the lane change maneuver. 

\begin{figure}[ht]
    \centering
    \includegraphics[trim={0.0cm 0.0cm 0.0cm 0.0cm},clip,width=8.4cm]{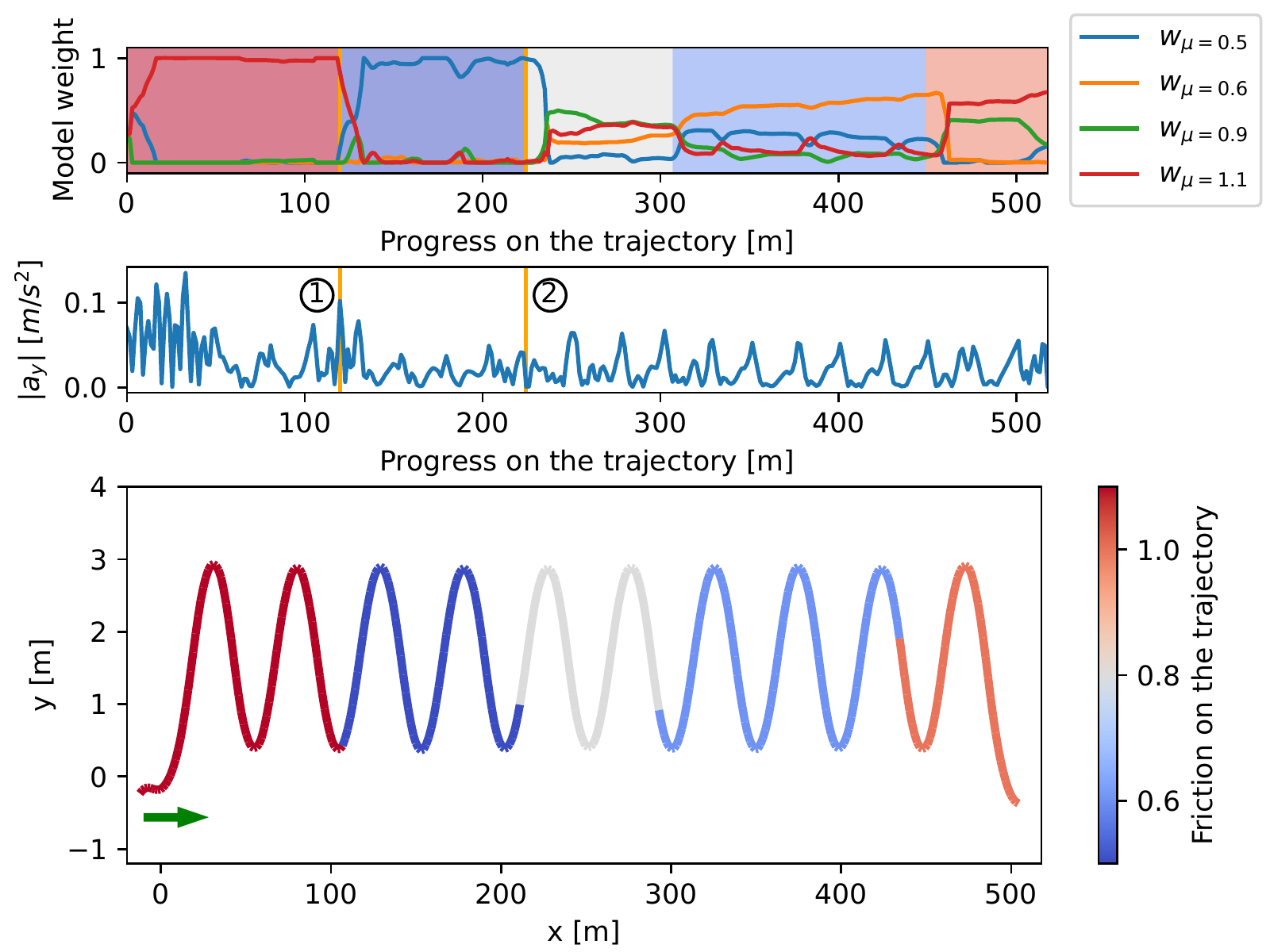}
    \vspace{-10pt}
    \caption{Reference trajectory with color-coded friction regions (Bottom), vehicle lateral acceleration in $\text{m}/\text{s}^2$ (Middle), and model weights used by the EGP-MPC (Top). Weights start changing faster when the system is excited (orange line 1) than when the system is not well excited (orange line 2).}
    \label{fig:N_models-reference}
\end{figure}

\textbf{Effect of System Excitation}
In this experiment we demonstrate how the EGP-MPC works with 4 models and how system excitation can affect weight transition. Models are created from data collected at friction coefficients $\theta^1 = 0.5$, $\theta^2 = 0.6$, $\theta^3 = 0.9$, and $\theta^4 = 1.1$. We use the dual lane change maneuver described in the previous experiment. The target velocity is set to $15 m/s$ and the maneuver is split into five segments of friction coefficients $[1.1, 0.5, 0.8, 0.6, 1.0]$. Friction segments and weight estimation by EGP-MPC algorithm are shown in Figure~\ref{fig:N_models-reference}. Locations of the first two friction transitions are specifically chosen to demonstrate how weight estimation depends on the system excitation. The first friction change is during the turn. This corresponds to the spike in the lateral acceleration (orange line 1, middle graph, Figure~\ref{fig:N_models-reference}), and immediate weight change (orange line 1, top graph, Figure~\ref{fig:N_models-reference}). The second friction transition is on the straight line, so the system is not well excited (orange line 2, middle graph, Figure~\ref{fig:N_models-reference}), and it takes much longer for the weight to start changing (orange line 2, top graph, Figure~\ref{fig:N_models-reference}).

\begin{figure}[ht]
    \centering
    \includegraphics[trim={0.0cm 0.0cm 0.0cm 0.0cm},clip,width=8.4cm]{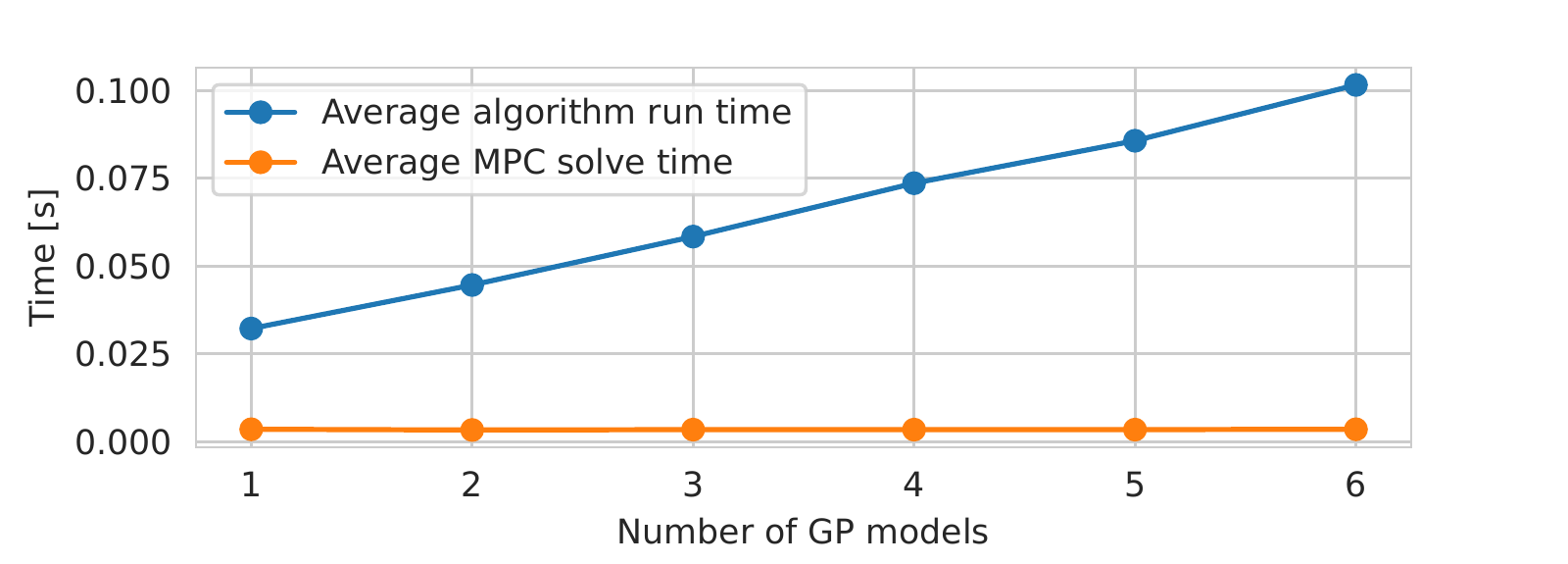}
    \vspace{-9pt}
    \caption{Scalability of the EGP-MPC algorithm with an increasing number of models. The average computation time of the EGP-MPC grows linearly only because we compute the linearization of models sequentially. It is possible to compute this linearization of models in parallel which would improve computation time scaling with increasing number of models.}
    \label{fig:time_scaling}
\end{figure}

\textbf{Computation time scaling}
Finally we discuss the runtime of the current implementation of the EGP-MPC algorithm. Because we are calculating the model ensemble before solving the MPC, the MPC solving time does not change with an increasing number of models (Figure~\ref{fig:time_scaling}). The current implementation calculates the linearization of models sequentially so the EGP-MPC computation time increases linearly (Figure~\ref{fig:time_scaling}). However, it is possible to compute this linearization in parallel which would improve computation time scaling with increasing number of models. 

\begin{figure}[ht]
    \centering
    \includegraphics[trim={0.0cm 0.0cm 0.0cm 0.0cm},clip,width=8.4cm]{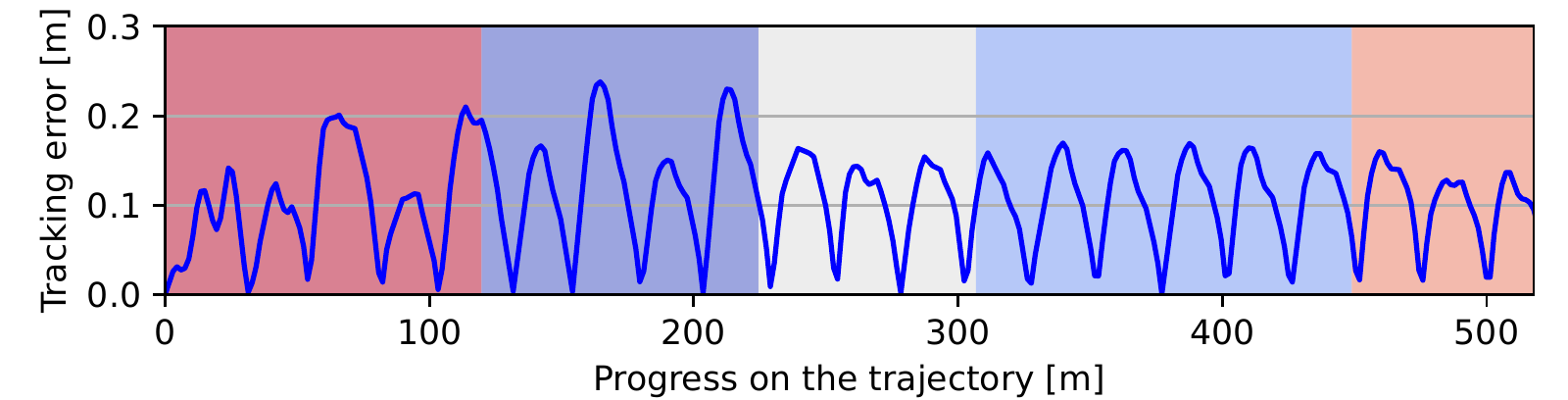}
    \vspace{-10pt}
    \caption{Tracking error while performing the dual lane change maneuver using four models.}
    \label{fig:N_models-tracking_error}
\end{figure}

\subsection{Limitations}
\vspace{-6pt}

We observe three main limitations of our approach. First, the estimation of the weight vector as a convex combination limits the space of unseen parameters that can be modelled to the convex combination of the trained parameters. Secondly, the algorithm scales linearly in terms of computation time as the number of GPs increases. This can be improved by linearizing the \acp{GP} in parallel through multi-threading or similar approaches. Finally, the weights estimated by the algorithm do not necessarily correlate with the actual value of the parameters of the system. This can be attributed to a lack of excitation of the system dynamics as can be seen when driving in straight lines at constant speed on the Sao Paolo racetrack (i.e Progress on the track 550m to 1000m).

\section{Conclusion}
\vspace{-6pt}
In this work, we presented EGP-MPC as an algorithm for driving on multiple friction surfaces. It leverages a library of pre-trained models, each corresponding to a different surface, to achieve higher state prediction accuracy. During the online update, it chooses the convex combination of models that best represents the surface that the car currently drives on. As shown in the results, the model combination clearly corresponds to different surface types. It is also shown that the algorithm behaves well on surfaces and trajectories that the algorithm was not trained on. With the use of \acp{GP}, it is also possible to reason about prediction uncertainty. For future work, EGP-MPC can be further constrained using the ensembled variance of the \acp{GP} to accommodate for this uncertainty.

\begin{ack}
\vspace{-6pt}
We would like to thank Mr. Johannes Betz from the University of Pennsylvania for his help with the initial formulation of the kinematic MPC.

\textit{This work was supported in part by NSF CCRI \#1925587 and DARPA \#FA8750-20-C-0542 (Systemic Generative Engineering). The views, opinions, and/or findings expressed are those of the author(s) and should not be interpreted as representing the official views or policies of the Department of Defense or the U.S. Government.}
\end{ack}

\vspace{-4pt}
\bibliography{ifacconf}             

\begin{thebibliography}{21}
\providecommand{\natexlab}[1]{#1}
\providecommand{\url}[1]{\texttt{#1}}
\providecommand{\urlprefix}{URL }
\expandafter\ifx\csname urlstyle\endcsname\relax
  \providecommand{\doi}[1]{doi:\discretionary{}{}{}#1}\else
  \providecommand{\doi}{doi:\discretionary{}{}{}\begingroup
  \urlstyle{rm}\Url}\fi

\bibitem[{Agrawal et~al.(2018)Agrawal, Verschueren, Diamond, and
  Boyd}]{agrawal2018rewriting}
Agrawal, A., Verschueren, R., Diamond, S., and Boyd, S. (2018).
\newblock A rewriting system for convex optimization problems.
\newblock \emph{Journal of Control and Decision}, 5(1), 42--60.

\bibitem[{Althoff et~al.(2017)Althoff, Koschi, and
  Manzinger}]{althoff_commonroad_2017}
Althoff, M., Koschi, M., and Manzinger, S. (2017).
\newblock {CommonRoad}: {Composable} benchmarks for motion planning on roads.
\newblock In \emph{2017 {IEEE} {Intelligent} {Vehicles} {Symposium} ({IV})},
  719--726. IEEE, Los Angeles, CA, USA.
\newblock \doi{10.1109/IVS.2017.7995802}.

\bibitem[{Betz et~al.(2022)Betz, Zheng, Liniger, Rosolia, Karle, Behl, Krovi,
  and Mangharam}]{Betz2022_RacingSurvey}
Betz, J., Zheng, H., Liniger, A., Rosolia, U., Karle, P., Behl, M., Krovi, V.,
  and Mangharam, R. (2022).
\newblock Autonomous vehicles on the edge: A survey on autonomous vehicle
  racing.
\newblock \emph{IEEE Open J. Intell. Transp. Syst.}, 3, 458--488.
\newblock \doi{10.1109/ojits.2022.3181510}.

\bibitem[{Christ et~al.(2021)Christ, Wischnewski, Heilmeier, and
  Lohmann}]{christ2021time}
Christ, F., Wischnewski, A., Heilmeier, A., and Lohmann, B. (2021).
\newblock Time-optimal trajectory planning for a race car considering variable
  tyre-road friction coefficients.
\newblock \emph{Vehicle system dynamics}, 59(4), 588--612.

\bibitem[{Diamond and Boyd(2016)}]{diamond2016cvxpy}
Diamond, S. and Boyd, S. (2016).
\newblock {CVXPY}: {A} {P}ython-embedded modeling language for convex
  optimization.
\newblock \emph{Journal of Machine Learning Research}, 17(83), 1--5.

\bibitem[{Gardner et~al.(2018)Gardner, Pleiss, Weinberger, Bindel, and
  Wilson}]{gardner_gpytorch_2021}
Gardner, J., Pleiss, G., Weinberger, K.Q., Bindel, D., and Wilson, A.G. (2018).
\newblock Gpytorch: Blackbox matrix-matrix gaussian process inference with gpu
  acceleration.
\newblock In S.~Bengio, H.~Wallach, H.~Larochelle, K.~Grauman, N.~Cesa-Bianchi,
  and R.~Garnett (eds.), \emph{Advances in Neural Information Processing
  Systems}, volume~31. Curran Associates, Inc.

\bibitem[{Hewing et~al.(2018)Hewing, Liniger, and
  Zeilinger}]{hewing2018cautious}
Hewing, L., Liniger, A., and Zeilinger, M.N. (2018).
\newblock Cautious nmpc with gaussian process dynamics for autonomous miniature
  race cars.
\newblock In \emph{2018 European Control Conference (ECC)}, 1341--1348. IEEE.

\bibitem[{Hewing et~al.(2020)Hewing, Wabersich, Menner, and
  Zeilinger}]{hewing2020learning}
Hewing, L., Wabersich, K.P., Menner, M., and Zeilinger, M.N. (2020).
\newblock Learning-based model predictive control: Toward safe learning in
  control.
\newblock \emph{Annual Review of Control, Robotics, and Autonomous Systems}, 3,
  269--296.

\bibitem[{ISO3888-2:2011(2011)}]{iso2011passenger}
ISO3888-2:2011 (2011).
\newblock Passenger cars--test track for a severe lane-change manoeuvre--part
  2: obstacle avoidance.

\bibitem[{Jain et~al.(2021)Jain, O'Kelly, Chaudhari, and
  Morari}]{jain_bayesrace_2020}
Jain, A., O'Kelly, M., Chaudhari, P., and Morari, M. (2021).
\newblock Bayesrace: Learning to race autonomously using prior experience.
\newblock In J.~Kober, F.~Ramos, and C.~Tomlin (eds.), \emph{Proceedings of the
  2020 Conference on Robot Learning}, volume 155 of \emph{Proceedings of
  Machine Learning Research}, 1918--1929. PMLR.

\bibitem[{Laurense et~al.(2017)Laurense, Goh, and Gerdes}]{laurense2017path}
Laurense, V.A., Goh, J.Y., and Gerdes, J.C. (2017).
\newblock Path-tracking for autonomous vehicles at the limit of friction.
\newblock In \emph{2017 American control conference (ACC)}, 5586--5591. IEEE.

\bibitem[{Le and Nghiem(2021)}]{le_receding_2021}
Le, V.A. and Nghiem, T.X. (2021).
\newblock A receding horizon approach for simultaneous active learning and
  control using gaussian processes.
\newblock In \emph{2021 IEEE Conference on Control Technology and Applications
  (CCTA)}, 453--458.
\newblock \doi{10.1109/CCTA48906.2021.9659046}.

\bibitem[{O’Kelly et~al.(2020)O’Kelly, Zheng, Karthik, and
  Mangharam}]{okelly_f1tenth_2020}
O’Kelly, M., Zheng, H., Karthik, D., and Mangharam, R. (2020).
\newblock {F1TENTH}: {An} {Open}-source {Evaluation} {Environment} for
  {Continuous} {Control} and {Reinforcement} {Learning}.
\newblock \emph{Proceedings of Machine Learning Research}, 13.

\bibitem[{Pleiss et~al.(2018)Pleiss, Gardner, Weinberger, and
  Wilson}]{noauthor_180306058_nodate}
Pleiss, G., Gardner, J., Weinberger, K., and Wilson, A.G. (2018).
\newblock Constant-time predictive distributions for {G}aussian processes.
\newblock In J.~Dy and A.~Krause (eds.), \emph{Proceedings of the 35th
  International Conference on Machine Learning}, volume~80 of \emph{Proceedings
  of Machine Learning Research}, 4114--4123. PMLR.

\bibitem[{Ragan-Kelley et~al.(2017)Ragan-Kelley, Adams, Sharlet, Barnes, Paris,
  Levoy, Amarasinghe, and Durand}]{ragan-kelley_halide_2017}
Ragan-Kelley, J., Adams, A., Sharlet, D., Barnes, C., Paris, S., Levoy, M.,
  Amarasinghe, S., and Durand, F. (2017).
\newblock Halide: decoupling algorithms from schedules for high-performance
  image processing.
\newblock \emph{Communications of the ACM}, 61(1), 106--115.
\newblock \doi{10.1145/3150211}.

\bibitem[{Rodriguez et~al.(2021)Rodriguez, Rosolia, Ames, and
  Yue}]{rodriguez_learning_2021}
Rodriguez, I.D.J., Rosolia, U., Ames, A.D., and Yue, Y. (2021).
\newblock Learning to control an unstable system with one minute of data:
  Leveraging gaussian process differentiation in predictive control.
\newblock In \emph{2021 IEEE/RSJ International Conference on Intelligent Robots
  and Systems (IROS)}, 3896–3903. IEEE Press.
\newblock \doi{10.1109/IROS51168.2021.9636786}.

\bibitem[{Rosolia and Borrelli(2020)}]{rosolia_learning_2019}
Rosolia, U. and Borrelli, F. (2020).
\newblock Learning how to autonomously race a car: A predictive control
  approach.
\newblock \emph{IEEE Transactions on Control Systems Technology}, 28(6),
  2713--2719.
\newblock \doi{10.1109/TCST.2019.2948135}.

\bibitem[{Rosolia et~al.(2018)Rosolia, Zhang, and Borrelli}]{rosolia2018data}
Rosolia, U., Zhang, X., and Borrelli, F. (2018).
\newblock Data-driven predictive control for autonomous systems.
\newblock \emph{Annual Review of Control, Robotics, and Autonomous Systems}, 1,
  259--286.

\bibitem[{Stellato et~al.(2020)Stellato, Banjac, Goulart, Bemporad, and
  Boyd}]{stellato_osqp_2020}
Stellato, B., Banjac, G., Goulart, P., Bemporad, A., and Boyd, S. (2020).
\newblock {OSQP}: an operator splitting solver for quadratic programs.
\newblock \emph{Mathematical Programming Computation}, 12(4), 637--672.
\newblock \doi{10.1007/s12532-020-00179-2}.

\bibitem[{Van~Gennip(2018)}]{van2018vehicle}
Van~Gennip, M. (2018).
\newblock \emph{Vehicle dynamic modelling and parameter identification for an
  autonomous vehicle}.
\newblock Master's thesis, University of Waterloo.

\bibitem[{Vaskov et~al.(2022)Vaskov, Quirynen, Menner, and
  Berntorp}]{vaskov_SNMPC_2022}
Vaskov, S., Quirynen, R., Menner, M., and Berntorp, K. (2022).
\newblock Friction-{Adaptive} {Stochastic} {Predictive} {Control} for
  {Trajectory} {Tracking} of {Autonomous} {Vehicles}.
\newblock In \emph{2022 {American} {Control} {Conference} ({ACC})}, 1970--1975.
  IEEE, Atlanta, GA, USA.
\newblock \doi{10.23919/ACC53348.2022.9867523}.

\end{thebibliography}

\end{document}